\documentclass{article}

\PassOptionsToPackage{numbers,square}{natbib}
 \usepackage[preprint]{neurips_2026}

\usepackage[utf8]{inputenc} % allow utf-8 input
\usepackage[T1]{fontenc}    % use 8-bit T1 fonts
\usepackage{hyperref}       % hyperlinks
\usepackage{url}            % simple URL typesetting
\usepackage{booktabs}       % professional-quality tables
\usepackage{amsfonts}       % blackboard math symbols
\usepackage{nicefrac}       % compact symbols for 1/2, etc.
\usepackage{microtype}      % microtypography
\usepackage{xcolor}         % colors
\usepackage{amsmath}
\usepackage{amsfonts}
\usepackage{algorithm}  
\usepackage{algpseudocode}
\usepackage{bbm}
\usepackage{multirow} % Add this package for multirow 
\usepackage{enumitem}
\setlist{leftmargin=0.2in}

\DeclareMathOperator*{\argmax}{arg\,max}
\newcommand{\bftab}{\fontseries{b}\selectfont}
\usepackage{adjustbox}
\usepackage{listings,booktabs}
\usepackage{cleveref}
\usepackage{todonotes}
\Crefname{appendix}{Appendix}{Appendices}
\usepackage{wrapfig}

\title{Memory Efficient Full-gradient Attacks (MEFA) Framework for Adversarial Defense Evaluations}

\author{%
  Yuan Du\\
  Department of Statistics and Data Science\\
  University of Central Florida\\
  Orlando, FL 32816, USA \\
  \texttt{yuan\_du@ucf.edu} \\
  \And
   Mitchel Hill \\
  InnoPeak Technology \\
  Palo Alto, CA 94303, USA \\
  \texttt{mitch.hill@innopeak.com } \\
  \AND
  HanQin Cai \\
  Department of Statistics and Data Science\\
  Department of Computer Science\\
  University of Central Florida\\
  Orlando, FL 32816, USA\\
  \texttt{hqcai@ucf.edu} \\
}

\begin{document}

\maketitle

\begin{abstract}
  This work studies the robust evaluation of iterative stochastic purification defenses under white-box adversarial attacks. Our key technical insight is that gradient checkpointing makes exact end-to-end gradient computation through long purification trajectories practical by trading additional recomputation for substantially lower memory usage. This enables full-gradient adaptive attacks against diffusion- and Langevin-based purification defenses, where prior evaluations often resort to approximate backpropagation due to memory constraints. These approximations can weaken the attack signal and risk overestimating robustness. In parallel, stochasticity in iterative purification is frequently under-controlled, even though different purification trajectories can substantially change reported robustness metrics. Building on this insight, we introduce a memory-efficient full-gradient evaluation framework for stochastic purification defenses. The framework combines checkpointed backpropagation with evaluation protocols that control stochastic variability, thereby reducing memory bottlenecks while preserving exact gradients. We evaluate diffusion-based purification and Langevin sampling with Energy-Based Models (EBMs), demonstrating that full-gradient attacks uncover vulnerabilities missed by approximate-gradient evaluations. Our framework yields stronger state-of-the-art $\ell_{\infty}$ and $\ell_{2}$ white-box attacks and further supports probing out-of-distribution robustness. Overall, our results show that exact-gradient evaluation is essential for reliable benchmarking of iterative stochastic defenses.
\end{abstract}

\section{Introduction}
\vspace{-0.12in}

Iterative stochastic purification defenses have emerged as a promising approach for adversarial robustness because they can be applied before classification without retraining the underlying classifier. Diffusion-based purification and Langevin sampling with Energy-Based Models (EBMs) are representative examples of this paradigm. However, their reliable white-box evaluation remains challenging: a strong adaptive attack must differentiate through a long sequence of stochastic purification steps, while naively storing all intermediate states during backpropagation incurs prohibitive memory cost. As a result, existing evaluations often rely on approximate gradients, adjoint-style methods, surrogate backward passes, or other memory-saving approximations. These approximations may weaken the attack signal and overestimate robustness.

The central insight of this work is that exact full-gradient attacks against iterative stochastic purification defenses can be made practical with memory-efficient checkpointed gradient computation. We propose \textbf{Memory-Efficient Full-gradient Attack} (\textbf{MEFA}), a framework that computes gradients through the full purification trajectory while reducing memory complexity with respect to the number of purification steps to $\mathcal{O}(1)$. MEFA detaches the forward purification trajectory and reconstructs the necessary local computation during backward propagation, enabling exact chain-rule gradient computation without storing the full computation graph. This makes strong adaptive attacks such as PGD with Expectation Over Transformation (PGD+EOT) \citep{athalye2018,aleks2017deep} feasible for iterative stochastic purification defenses under realistic memory constraints.

This evaluation problem is important because deep neural networks (DNNs), despite achieving state-of-the-art performance in vision \citep{krizhevsky2012imagenet,dosovitskiy2020image,liu2021swin}, language \citep{vaswani2017attention,touvron2023llama,openai2024gpt4technicalreport}, and audio tasks \citep{gong2022ast,chen2022beats}, remain vulnerable to adversarial examples. Small, human-imperceptible perturbations can cause high-confidence misclassification \citep{szegedy2013intriguing,IanGan2014}. For purification-based defenses, robustness evaluation must include not only the final classifier, but also the full randomized iterative preprocessing mechanism. Otherwise, the reported robustness may reflect an incomplete white-box threat model rather than true defensive strength.

Prior work illustrates the difficulty of this evaluation. DiffPure \citep{nie2022diffusion} uses diffusion purification to remove adversarial perturbations and adopts the adjoint method to approximate gradients through the reverse Stochastic Differential Equation (SDE) with low memory cost. However, these gradients are approximate rather than exact full gradients through the purification trajectory. Subsequent studies have shown that diffusion-based defenses can lose substantial robustness under stronger adaptive or surrogate attacks \citep{lee2023robust,kang2024diffattack}, suggesting that gradient quality is central to reliable robustness evaluation. Similar challenges arise in Langevin-based EBM purification, where the defense also relies on iterative stochastic sampling. Without exact-gradient attacks, reported robustness may reflect limitations of the evaluation procedure rather than genuine defensive strength.

A second challenge is stochasticity. Even if exact gradients are available for a fixed purification trajectory, a stochastic purification defense does not define a single deterministic prediction for each input. Instead, it induces a distribution over purified samples and classifier outputs. Reported robustness can therefore depend on protocol choices such as the number of purification replicates, the number of adversarial candidates generated per input, and whether the final PGD iterate or the highest-loss iterate is selected. Too few purification samples can produce high-variance results, while validating many adversarial candidates with insufficient stochastic replicates can distort attack success estimates.

MEFA addresses both challenges: memory-efficient exact-gradient attack generation and robust stochastic validation. In the attack process, MEFA computes exact full gradients through iterative purification with $\mathcal{O}(1)$ memory complexity in the number of purification steps, enabling PGD+EOT attacks to optimize perturbations using full adaptive gradients rather than weakened approximations. In the validation process, MEFA treats the stochastic defense as an underlying deterministic classifier defined by the expectation over infinitely many purification trajectories. Since this expectation cannot be computed exactly, MEFA approximates it using sufficient parallel purification replicates and determines the number of replicates needed to obtain stable, nearly deterministic predictions. Together, these components reduce both gradient-induced and randomness-induced evaluation artifacts.

Our key contributions are:
\vspace{-0.12in}
\begin{itemize}
    \item \textbf{A Memory Efficient Full Gradient Attack via Chain Rule}: We introduce a framework for attacking iterative purification defenses using strong gradient checkpointing adaptive attacks, overcoming prohibitive memory costs associated with full gradient computation during backpropagation in the adversarial attack process. See \Cref{sec:attack_process}

\vspace{-0.05in}
    \item \textbf{Robust Validation}: We formalize stochastic purification as an expected classifier over purification randomness and provide a validation process that uses sufficient parallel replicates to obtain stable robustness estimates. See \Cref{sec:validation_process}.

\vspace{-0.05in}
    \item \textbf{SOTA Attack Success Rate against Iterative Stochastic Defenses}: Using MEFA, we expose critical limitations of score SDE-, DDPM-, and EBM-based purification defenses and achieve stronger state-of-the-art $\ell_{\infty}$ and $\ell_{2}$ white-box attacks. See \Cref{sec:exp}. 

\vspace{-0.05in}
    \item \textbf{Insights on OOD Robustness}:  Furthermore, the diffusion model shows the potential to defend OOD datasets. However, the performance is limited to the data distribution similarity between the pre-trained and defended datasets. (See \Cref{sec:exp})
\end{itemize}
\vspace{-0.1in}

\vspace{-0.08in}
\section{Preliminaries}
\vspace{-0.08in}
\noindent\textbf{Diffusion Models.} 
Diffusion models consist of two processes: first, a forward process that gradually adds noise to the original input until it becomes random noise, and second, a reverse process that generatively recovers the original input by removing the predicted noise at each time step. There are two types of diffusion models: discrete Diffusion Model (DDPM-based) and continuous diffusion model (score SDE-based).  

\vspace{0.1in}
\textit{Discrete Diffusion Model.}
Given the input sample as $\mathbf{x}_0 \sim q(\mathbf{x})$, the forward process adds Gaussian noise over $T$ steps via a variance schedule $\{\beta_t \in (0,1)\}_{t=1}^{T}$ with $q(\mathbf{x}_{t}|\mathbf{x}_{t-1})=\mathcal{N}(\mathbf{x}_{t};\sqrt{1-\beta_{t}}\mathbf{x}_{t-1},\beta_{t}I)$ and $q(\mathbf{x}_{1:T}|\mathbf{x}_{0})=\prod_{t=1}^{T}q(\mathbf{x}_{t}|\mathbf{x}_{t-1})$. The reverse process learns $p_{\theta}(\mathbf{x}_{t-1} \vert \mathbf{x}_t) = \mathcal{N}(\mathbf{x}_{t-1}; \boldsymbol{\mu}_\theta(\mathbf{x}_t, t), \boldsymbol{\Sigma}_\theta(\mathbf{x}_t, t))$ by optimizing the simple loss:
\begin{equation}
\mathcal{L}(\theta) = \mathbb{E}_{\mathbf{x}_0,\varepsilon, t \sim U[1,T]} \left[ \parallel \varepsilon - \hat{\varepsilon}_{\theta} (\mathbf{x}_t, t)\parallel ^2 \right].
\label{eqn:sde_loss}
\end{equation}

Denoising Diffusion Probabilistic Model (DDPM) \citep{ho2020denoising} employs the update:
\begin{equation}
\mathbf{x}_{t-1} = \frac{1}{\sqrt{\alpha_t}} \Big( \mathbf{x}_t - \frac{1 - \alpha_t}{\sqrt{1 - \bar{\alpha}_t}} \varepsilon_{\theta}(\mathbf{x}_t, t) \Big) + \sigma_t \mathbf{z},
\label{eqn:ddpm}
\end{equation}
where $\alpha_{t}=1-\beta_{t}, \bar{\alpha}_{t}=\prod_{t=1}^{T}\alpha_{t}$, $\sigma_{t} = \sqrt{ \beta_{t} (1 - \bar{\alpha}_{t-1}) / (1 - \bar{\alpha}_{t}) }$, $\mathbf{z} \sim \mathcal{N}(0,I)$.

\vspace{0.1in}
\textit{Continuous Diffusion Model.} 
The forward process is a Stochastic Differential Equation (SDE):$d\mathbf{x} = \mu(\mathbf{x},t)dt + \sigma(\mathbf{x},t)dw_t$ with drift $\mu(\mathbf{x},t)$, volatility $\sigma(\mathbf{x},t)$, and Wiener process $w_t$.The reverse-time SDE \citep{song2020score} is:  
\begin{equation}
d\mathbf{x} = [\mu(\mathbf{x},t) - \sigma(\mathbf{x},t)^2 \nabla_{\mathbf{x}} \log p(\mathbf{x}, t) ]dt + \sigma(\mathbf{x},t)d\bar{w_t},
\label{eqn:rdiff}
\end{equation}
where a scoring network $s_{\theta} (\mathbf{x},t)$ estimates $\nabla_{\mathbf{x}} \log p(\mathbf{x}, t)$. Training minimizes the weighted score matching loss: 
\begin{equation}
\mathcal{L}(\theta) = \min_{\theta}\int_{t=0}^{1} \mathbb{E}_{ {p(\mathbf{x})p_{0t}(\hat{\mathbf{x}}(t)|x(0))} }\left[ \lambda_t \left\Vert\nabla_{\hat{\mathbf{x}}} \log (p_{0t}(\hat{\mathbf{x}}(t)|\mathbf{x}(0))) - s_{\theta}(\hat{\mathbf{x}}(t), t) \right\Vert^2_2 \right],
\label{eqn:ldiff}
\end{equation}
where $\lambda_t$ is the time-dependent weighting coefficient \citep{song2020sliced, vincent2011connection}. 
% $P_{0t}(\hat{\mathbf{x}_t}|x_0)$ is the transition probability from $\mathbf{x}_0$ to $\mathbf{x}_t$. 

% \vspace{-0.001in}
\textbf{Energy Based Models.} 
EBM is a Gibbs-Boltzmann density \citep{xie2016theory}. A deep EBM has the form: 
\begin{equation}
P(\mathbf{x}; \theta)=\frac{1}{Z(\theta)} \exp\{-U(\mathbf{x};\theta)\}, \label{eqn:ebm}
\end{equation}
where the energy $U(\mathbf{x};\theta)$ is a ConvNet with a scalar output. The intractable $Z(\theta) = \int_{\mathcal{X}} \exp{[-U(\mathbf{x};\theta)]}\,d\mathbf{x}$ is circumvented via \textit{Markov Chain Monte Carlo} (MCMC) sampling in unsupervised \textit{Maximum Likelihood Estimation} (MLE) learning.
The gradient used to learn $\theta$ is approximated:
\begin{equation}
\begin{aligned}
  \nabla \mathcal{L}(\theta) &= \nabla  \mathbb{E}_q [ U(\mathbf{x};\theta) ]  - \nabla \mathbb{E}_{P_{\theta}} [ U(\mathbf{x};\theta) ] \\
 & \approx \frac{1}{n}\sum_{i=1}^n \nabla_\theta U(\mathbf{x}_i^+;\theta)-\frac{1}{m}\sum_{i=1}^{m} \nabla_\theta U(\mathbf{x}_i^-;\theta),
\end{aligned}
  \label{eqn:mlobject}
  \end{equation}
where $\{\mathbf{x}_i^+\}_{i=1}^n$ and $\{\mathbf{x}_i^-\}_{i=1}^m$ are positive samples and negative samples from model $P(\mathbf{x};\theta)$. In practice, positive samples $\{\mathbf{x}_i^+\}_{i=1}^n$ are a batch of training images and negative samples $\{\mathbf{x}_i^-\}_{i=1}^m$ are obtained from $K$-step Langevin Dynamics (LD):
\begin{equation}
\mathbf{x}^{(k+1)} = \mathbf{x}^{(k)} - \frac{\eta^2}{2} \nabla_{\mathbf{x}^{(k)}} U(\mathbf{x}^{(k)} ;\theta) + \eta \mathbf{z}_k , \label{eqn:langevin}
\end{equation}
% \todo{use a different letter for step size}
with step size $\eta$ and $\mathbf{z}_k \sim \mathcal{N}(0, I^D)$. Initialization of Langevin chains $\{\mathbf{x}^-_{i, 0}\}_{i=1}^n$ critically influences model behavior.

\vspace{-0.1in}
\section{Memory Efficient Full-gradient Attack (MEFA) Framework}
\vspace{-0.15in}

\begin{figure*}[ht]
    \centering
    \includegraphics[width=.86\textwidth]{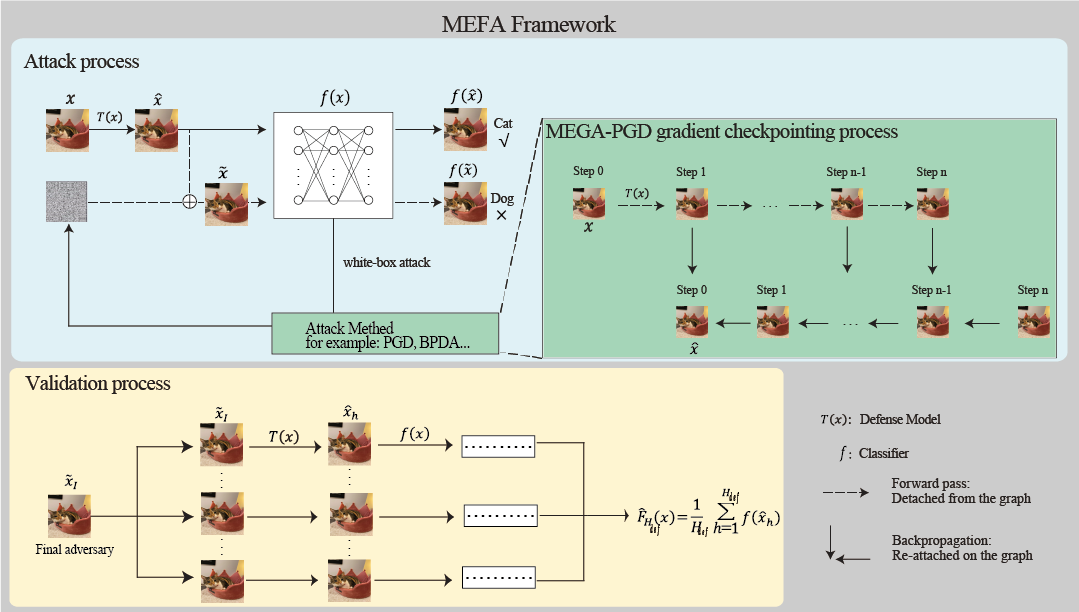}
    \vspace{-0.08in}
    \caption{\small Diagram for adversarial attack process (top) and validation process (bottom). In the MEFA-PGD gradient checkpointing process (top right): each image state represents the forward computation steps from initial state $\mathbf{x}$ to the final state $\hat{\mathbf{x}}$ through attack on the defense model transformation $T(\mathbf{x})$, specifically described as from $\mathbf{x}^{(0)}$ to $\mathbf{x}^{(K)}$ for EBM in \eqref{eqn:langevin},  $\mathbf{x}_T$ to $\mathbf{x}_0$ for Diffusion model in \eqref{eqn:ddpm} or \eqref{eqn:rdiff}. All intermediate $\mathbf{x}$'s are detached and saved to CPU from the graph during the forward pass. In the backpropagation, each saved image state is re-attached to the graph and the gradient is computed with respect to the image state during the backward pass and passed to the next layer via the chain rule. In the defense validation process, the final adversarial state could be repeated with a number of replicates $H_\mathrm{d}$, go through the purification defense model $T(\mathbf{x})$ to check the efficacy of the attack and defense strength without intensive computing memory and time.}
    \label{fig:MEFA_framework}
    \vspace{-0.1in}
\end{figure*}

In this work, our objective is to tackle the difficulties of evaluating adversarial defenses based on iterative sampling. Adaptive adversarial attack on iterative stochastic purification defense is challenging due to 1) high memory and long time requirement for gradient calculation for the long sampling process, 2) stochastic transformation that induces gradient instability for the classifier and leads to high outcome variance that needs further analysis. In this section, we first formulate the adversarial attack against iterative stochastic purification, and then MEFA Framework is presented. Our methodological framework contains two main processes: Attack process and validation process. The adversarial attack process constitutes the initial memory-efficient computational process, while the validation process operates as an independent module for robustness validation to prevent overstatement of attack success. To optimize experimental reproducibility and computational efficiency, we advocate removing any stochasticity by defining the expectation of prediction over stochastic transformation, using many parallel purifications in a decoupled process in which researchers save adversarially perturbed purified images during the attack process for further analysis. Additional analyses can then leverage the final saved images to be re-evaluated by the validation pipeline. This validation process is flexible for comprehensive defense performance analysis without re-executing resource-intensive adversarial attack iterations. This MEFA Framework substantially reduces memory overhead and time cost during repeated evaluations. See \Cref{fig:MEFA_framework} for a pictorial illustration. 

\textbf{Formulation.}\label{sec:attack_formulation}
We first formulate the adversarial attack against the stochastic preprocessing transformation for diffusion-based and EBM-based defenses. Let $T(\mathbf{x})$ be the purification transformation with diffusion-based defenses, \eqref{eqn:ddpm} or \eqref{eqn:rdiff}, or EBM-based defenses \eqref{eqn:langevin} for an image state $\mathbf{x}\in \mathbb{R}^D$. The classifier $f$ defended by the transformation $T(\mathbf{x})$ is defined by $f(T(\mathbf{x}))$. Logits are presented as $f(\mathbf{x})$, and expected logits for EOT attack are presented as $F(\mathbf{x})$: 
\begin{equation}
F(\mathbf{x})=\mathbb{E}_{T(\mathbf{x})} [f(T(\mathbf{x}))]. 
\label{eqn:predict_label}
\end{equation}

\vspace{0.1in}
We can compute the EOT attack gradient as the average of finite samples according to the law of large numbers, where $H$ is the number of adversarial samples practically with $\nabla \mathbb{E}_{T(\mathbf{x})} [ f(T(\mathbf{x}))] \approx \nabla \frac{1}{H} \sum_{h=1}^H f(\hat{\mathbf{x}_h})$, where $\hat{\mathbf{x}_h} \sim T(\mathbf{x})$. In the attack, $H_{\mathrm{adv}}$ and $H_{\mathrm{d}}$ are denoted as the number of attack replicates and the number of defense replicates, respectively.

Given an input pair $(\mathbf{x}, y)$, an adversarial example $\tilde{\mathbf{x}}$ satisfies the following:
% \todo{missing tilde in x}
\begin{equation*}
\begin{aligned} 
& \textstyle{\argmax_{\tilde{\mathbf{x}}}}~~\mathbb{E}_{T(\tilde{\mathbf{x}})} [f(T(\tilde{\mathbf{x}}))] \neq y \\
& \text{subject to }~ \mathbb{E}_{T(\mathbf{x}),T(\tilde{\mathbf{x}})}[d(\mathbf{x},\tilde{\mathbf{x}})] < \varepsilon, \quad \mathbf{x} \in [0,1]^D.
\end{aligned}
\label{eqn:adv_x}
\end{equation*}
Let $\Pi_{\mathbf{x}+S}$ denote the projection on $\mathbf{x}+S$, where $S$ is the intersection of the image $\mathbf{x}\in [0, 1]^D$ and the $\ell_p \text{-norm}$ around image $\mathbf{x}$ for a small $\varepsilon >0$. The PGD attack  is iteratively updated through:
\begin{equation}
    \tilde{\mathbf{x}}^{j+1} = \Pi_{\mathbf{x}+S} (\tilde{\mathbf{x}}^j + \eta \text{ sign} \nabla_x^j \mathcal{L}( (F \circ T) (\mathbf{x}), y)|_{\mathbf{x}=\tilde{\mathbf{x}}_j} ).
\label{eqn:PGD}
\end{equation}
However, to calculate the full gradient $\nabla_\mathbf{x}^j \mathcal{L}( (F \circ T) (\mathbf{x}), y)|_{\mathbf{x}=\tilde{\mathbf{x}}_j}$ through $T(\mathbf{x})$ sampling process on the computational graph is known to be memory intensive. Alternative attack methods use a limited number of denoising steps to attack diffusion-based models \citep{nie2022diffusion, blau2022threatmodelagnosticadversarialdefense, lee2023robust, chen2023robust} due to the memory constraint. DiffAttack \citep{kang2024diffattack} uses segment-wise forwarding-backwarding algorithm to solve the memory problem, but formulates a surrogate loss, termed deviated-reconstruction loss (mse+ce) directly to the adversarial image state resulting in a weaker attack (see \Cref{tab:table1}). Other alternative attack methods that use fewer sampling steps, or gradient approximation, such as BPDA, report false security in EBM-based defenses \citep{du2019implicit,grathwohl2019modeling,yoon2021adversarial}.

\vspace{-0.08in}
\textbf{Attack Process.}\label{sec:attack_process}
Our attack process is a memory-efficient adaptive full gradient white-box adversarial attack, which has  stronger attack success rates than state-of-the-art alternative attack methods. The main components are based on iterative PGD attack and gradient checkpointing. In the PGD attack, we bound the perturbation in the $\ell_p$-norm epsilon ball. and we use adaptive or fixed adversarial steps.  Many researchers shy away from adversarial purification by iterative sampling-based defense for white-box attacks due to memory limitations. Alternative attack methods and gradient approximation methods have been developed, such as BPDA attack \citep{athalye2018obfuscated}, joint attack \citep{yoon2021adversarial}, adjoint attack \citep{nie2022diffusion}, surrogate attack \citep{chen2023robust}, DiffAttack \citep{kang2024diffattack} to overcome memory problems and obfuscated gradients. They all have shown very weak attack results (see \Cref{tab:table1} for DiffAttack, \Cref{tab:table3} for BPDA). Gradient checkpointing operates on the principle of recomputation during backpropagation to reduce memory consumption during backpropagation. Taking diffusion-based purification $D(\mathbf{x};\theta)$ as an example, during the forward purification pass, we store intermediate adversarial images $\mathbf{x}_T,\mathbf{x}_{T-1},\mathbf{x}_0$ and noises $\mathbf{z}_T,\mathbf{z}_{T-1},\mathbf{z}_0$, detached from the graph. During backpropagation, we put each of the intermediate adversarial images stored $\mathbf{x}_T,\mathbf{x}_{T-1},\mathbf{x}_0$ back on the graph, recompute the next state, and calculate the gradient of the recomputed state with respect to the current state iteratively until the final adversarial state to update the gradient via chain rule. Thus, the memory complexity is $\mathcal{O}(1)$ instead of $\mathcal{O}(T)$. 
We denote the input as $\mathbf{x}_t$ and the purified input as $\mathbf{x}'_t$. Then each gradient with respect to the input is as during iterative backpropagation at step $t \in [0,T]$ is
\begin{equation}
\frac{\partial \mathcal{L}}{\partial \mathbf{x}'_t}  =  \frac{\partial L}{\partial \mathbf{x}'_{t-1}}  \frac{\partial \mathbf{x}'_{t-1}}{\partial \mathbf{x}'_{t}}  =  \frac{\partial \mathcal{L}}{\partial \mathbf{x}'_{t-1}} \frac{\partial D(\mathbf{x}'_t, \mathbf{z}_t)}{\partial \mathbf{x}'_{t}}.     
\end{equation}
At $t$-th step, we recompute the next state $ \mathbf{x}'_{t-1} = D(\mathbf{x}'_t, \mathbf{z}_t)$ using the intermediate adversarial images stored, and calculate the gradient $\frac{\partial D(\mathbf{x}'_t, \mathbf{z}_t)}{\partial \mathbf{x}'_{t}}$ using chain rule to get $\frac{\partial L}{\partial \mathbf{x}_0}$. 

We summarize the pseudo-codes of the MEFA Framework PGD attack in \Cref{sec:FullGradAttack}, the gradient checkpointing algorithm in \Cref{sec:GradientCheckpointing}, and the comparison between gradient checkpointing and the segment-wise forwarding-backwarding algorithm \citep{kang2024diffattack} in \Cref{sec:algorithm_comparison}. 

\textbf{Why Full Gradients?}
Reliable robustness evaluation requires gradients that faithfully reflect the defended model. Adversarial examples exploit the local sensitivity of neural networks along gradient directions, so approximate gradients can misalign the attack optimization and underestimate vulnerability \citep{goodfellow2014explaining}. This has been a recurring failure mode in defenses that appeared robust under approximate or obfuscated gradients but were later broken by adaptive attacks with correct gradient information \citep{athalye2018obfuscated}. Exact, or provably equivalent, gradients also make evaluations less dependent on implementation-specific approximations, improving reproducibility and comparability across defenses. Accordingly, standard robustness evaluation guidelines emphasize the use of exact gradients whenever possible \citep{carlini2019evaluating}.

For stochastic purification defenses, the choice of adversarial state used for reporting adversarial accuracy also matters. prior works implement exact backpropagation \citep{kang2024diffattack} via checkpointing \citep{chen2016training}. We discover and fix issues in 
implementations and utilize a validation process to efficiently validate the saved adversarial states other than reporting attack success on the fly from the first broken stochastic state which can overestimate attack effectiveness. This decouples expensive gradient-based attack optimization from stochastic robustness estimation and enables controlled evaluation with parallel purifications. See \Cref{sec:experiment} for details.

\begin{table}[t]
\caption{Defense against APGD+EOT20 $\ell_\infty$ attacks ($\varepsilon=8/255$) from WRN-28-10 on CIFAR-10 compared with DiffAttack \citep{kang2024diffattack}.}
\label{tab:table1}
\vspace{-0.005in}
\centering
\scriptsize
\renewcommand{\arraystretch}{1.2} % Adjust row height
\setlength{\tabcolsep}{6pt} % Reduce column padding
\resizebox{\textwidth}{!}{ % Resize to fit one column
\begin{tabular}{lccc}
\toprule
\textbf{Defense Model} & \textbf{Attack Method} & \textbf{Nat. Accuracy (NA)} & \textbf{Adv. Accuracy (AA)} \\
\midrule
\multirow{6}{*}{\begin{tabular}[c]{@{}l@{}}Score SDE VP-SDE\\(Contin. Diffusion)\end{tabular}} 
& \textbf{Ours}, $H_{\mathrm{d}}=1$  & 89.00\% & 45.69\% \\
& \textbf{Ours}, $H_{\mathrm{d}}=50$ & 93.33\% & \bftab{42.75\%} \\ \cline{2-4}
& DiffAttack*, $H_{\mathrm{d}}=1$  & 89.00\% & 55.49\% \\
& DiffAttack*, $H_{\mathrm{d}}=50$  & 93.33\% & 61.57\% \\ \cline{2-4}
& DiffAttack** (Surrogate Attack \citep{kang2024diffattack}) & 87.50\% & 46.88\% \\ 
& DiffPure** (Adjoint Attack \citep{nie2022diffusion}) & 89.02\% & 70.64\% \\
\midrule
\multirow{5}{*}{\begin{tabular}[c]{@{}l@{}}2DUnet DDPM\\(Discrete Diffusion)\end{tabular}}
& \textbf{Ours}, $H_{\mathrm{d}}=1$  & 85.69\% & 40.20\% \\
& \textbf{Ours}, $H_{\mathrm{d}}=50$  & 92.94\% & \bftab{35.29\%} \\ \cline{2-4}
& DiffAttack*, $H_{\mathrm{d}}=1$ & 85.68\% & 64.70\% \\
& DiffAttack*, $H_{\mathrm{d}}=50$ & 89.80\% & 67.65\% \\ \cline{2-4}
& DiffAttack** (Surrogate Attack\citep{kang2024diffattack}) & 100\% & 54.69\% \\ 
\midrule
\multirow{2}{*}{Smooth EBM}
&  \textbf{Ours}, $H_{\mathrm{d}}=1$  & 72.16\% & 7.84\% \\
&  \textbf{Ours}, $H_{\mathrm{d}}=50$ & 85.69\% & \bftab{6.08\%} \\
\bottomrule
\end{tabular}
}
\vspace{0.5mm}
\vspace{-0.1in}
\begin{minipage}{\textwidth}
\scriptsize
\emph{Note:} *validated using our validation process in DiffAttack's pipeline, ** reported from the original paper. 
\vspace{-0.1in}
\end{minipage}
\vspace{-0.15in}
\end{table}

\textbf{Validation Process.}\label{sec:validation_process}
Although MEFA removes the memory bottleneck for full-gradient PGD attacks, attack generation remains computationally expensive. PGD with 100 attack steps can take several days for EBM- and score SDE-based defenses, even without parallel purification trials. At the same time, validating stochastic defenses with a single purification produces high-variance robustness estimates, while rerunning the full attack for different numbers of purification samples is impractical.

We therefore decouple attack generation from robustness estimation. The attack process saves the final adversarial states, and a separate validation process reuses these saved examples with a flexible number of defense replicates $H_d$. This enables efficient estimation of the expected stochastic defended classifier without rerunning attack optimization, and also supports variance analysis and visualization under different validation settings.

\begin{figure}[ht]
    \centering
    \includegraphics[width=.75\textwidth]{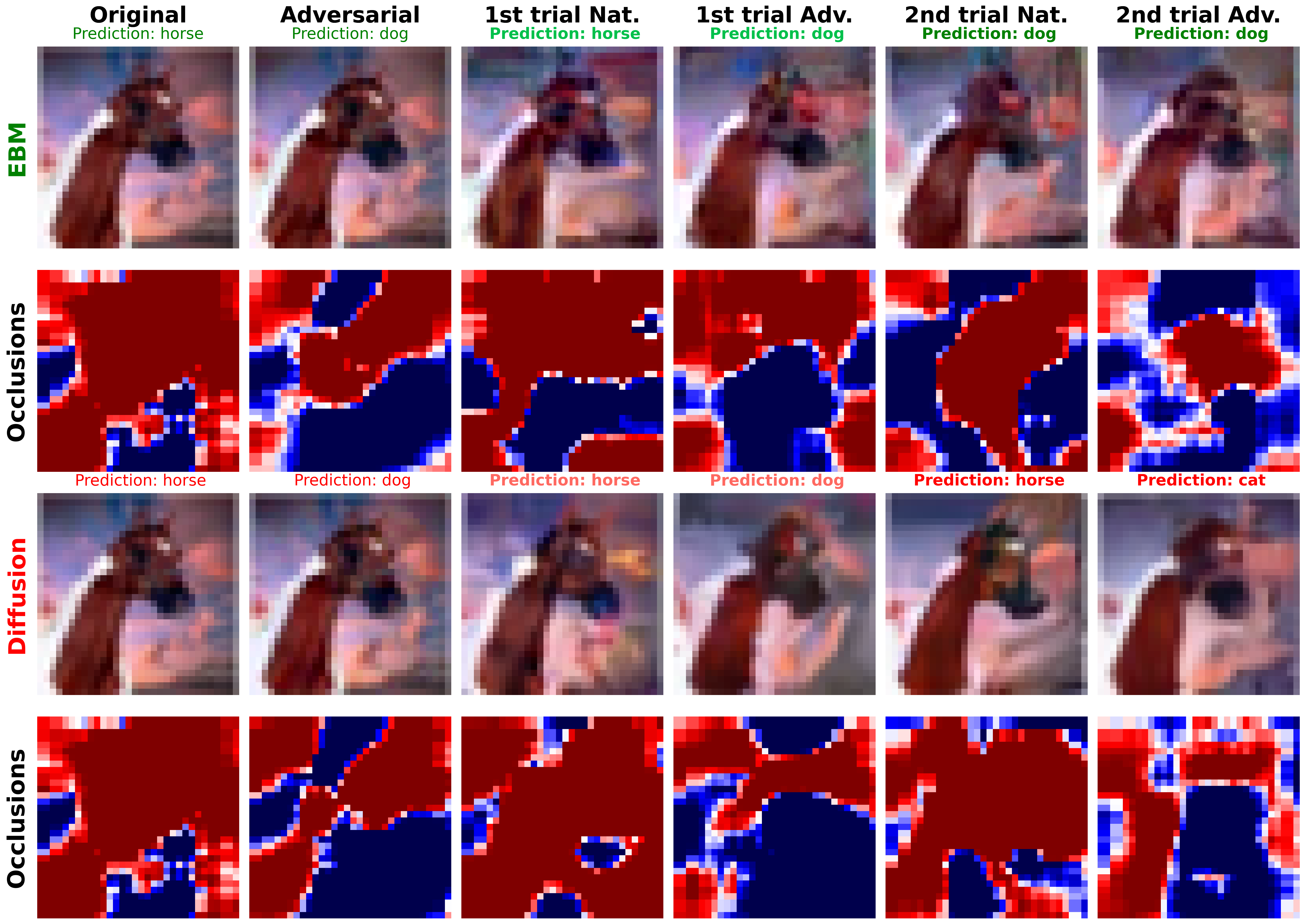}
    \vspace{-0.12in}
    \caption{\small Purified images present random prediction with occlusion attributions \citep{zeiler2013} where red and blue regions indicate high and low relevance areas to confidence reduction, respectively. \textit{Left to right}: original, adversarial state, 1st-trial natural state, 1st-trial adversarial state, 2nd-trial natural state, and 2nd-trial adversarial state. }
    \label{fig:image_states_random}
    \vspace{-0.05in}
\end{figure}

Empirically, single-purification validation can be highly unstable, even with EOT20 during attack generation, as shown in \Cref{fig:image_states_random}. Increasing $H_d$ reduces variance and stabilizes predictions, particularly for borderline cases. Larger $H_d$ amplifies the majority prediction: examples correctly classified by most purification trials become more reliably correct, while examples misclassified by most trials become more reliably incorrect. \Cref{tab:table6} confirms that higher $H_d$ reduces outcome variance.

The validation process is also substantially faster than full re-evaluation. For the EBM defense, validation takes 27 seconds per image, compared with 13 minutes per image for standard evaluation, giving over $30\times$ speedup. Because it operates directly on saved adversarial examples, the same pipeline can also serve as a post-hoc verification tool for existing attacks. Applying it to adversarial examples from DiffAttack \citep{kang2024diffattack}, we observe substantially higher robust accuracy after re-validation than the originally reported attack-stage results, indicating that stochastic validation protocol choices can materially affect robustness estimates.

In our experiments, $H_d=50$ provides a practical stability and cost trade-off on CIFAR-10. We therefore report both single-purification and $H_d=50$ validation results in the remaining experiments. See more in \Cref{sec:hdef_estimate}.

\begin{table}[t]
\centering
\caption{\small Defense against PGD+EOT20 $\ell_\infty$ attacks from WRN-20-10 on CIFAR-10 with the same 60 images.}
\label{tab:table6}
\vspace{-0.1in}
\resizebox{\textwidth}{!}{ % Resize table to fit page width
\begin{small}
\begin{tabular}{p{2cm}ccccccc}
\toprule
\multirow{2}{4cm}{{\scriptsize\textbf{Defense}}} 
& \multirow{2}{*}{{\scriptsize\textbf{Trial}}} 
& \multicolumn{2}{c}{{\scriptsize\textbf{$H_{\mathrm{d}}=1$}}} 
& \multicolumn{2}{c}{{\scriptsize\textbf{$H_{\mathrm{d}}=30$}}} 
& \multicolumn{2}{c}{{\scriptsize\textbf{$H_{\mathrm{d}}=50$}}}  \\
\cmidrule(lr){3-4} \cmidrule(lr){5-6} \cmidrule(lr){7-8}  
{\scriptsize\textbf{Model}} 
& {\scriptsize\textbf{No.~\,}} 
& {\scriptsize\textbf{NA}} 
& {\scriptsize\textbf{AA}}  
& {\scriptsize\textbf{NA}} 
& {\scriptsize\textbf{AA}} 
& {\scriptsize\textbf{NA}} 
& {\scriptsize\textbf{AA}}  \\
\midrule

\multirow{3}{3cm}{Score SDE VP-SDE \\ (Contin. Diffusion)} 
& 1 & 90.00\% & 48.33\% & 90.00\% & 55.00\% & 90.00\% & 55.00\% \\
& 2 & 88.33\% & 56.67\% & 90.00\% & 58.33\% & 90.00\% & 56.67\% \\
& 3 & 88.33\% & 58.33\% & 90.00\% & 60.00\% & 90.00\% & 58.33\% \\ 
\hline

\multirow{2}{4cm}{Mean \\ $\pm$ std} 
& & 88.89\% & 54.44\% & 90.00\% & 57.78\% & 90.00\% & 56.67\%\\
& & $\pm$0.96\% & $\pm$5.36\% & $\pm$0.00\%\, & $\pm$2.55\% & $\pm$0.00\%\, & $\pm$1.67\%\\

\bottomrule
\end{tabular}
\end{small}
}
\vspace{-0.15in}
\end{table}

\noindent\textbf{Limitations.} 
While the MEFA Framework enables a robust and memory-efficient evaluation of stochastic purification defenses, several directions remain open for future work. These include reducing the runtime overhead introduced by gradient checkpointing, extending MEFA to larger-scale models and datasets, and generalizing the validation replicate count across diverse settings. Additionally, future work will be needed for defense models with non-smooth activations. These extensions represent promising avenues to further strengthen evaluations of iterative stochastic purification defenses.

\vspace{-0.12in}
\section{Empirical Experiments}\label{sec:exp}
\vspace{-0.12in}
We compare EBM- and diffusion-based purification defenses under MEFA, including DDPM \citep{blau2022threatmodelagnosticadversarialdefense} and score SDE defenses \citep{nie2022diffusion}, across multiple white-box attacks. We focus in particular on DiffPure \citep{nie2022diffusion}, which evaluates score SDE purification using the adjoint method against $\ell_{\infty}$ APGD attacks with $\varepsilon=8/255$ on CIFAR-10 and $\varepsilon=4/255$ on ImageNet. For validation, we save the loss-optimized adversarial states for APGD attacks and the final adversarial states for PGD attacks. Since the adjoint method is introduced to address memory constraints, the attack can only access gradients of the final sampling state rather than exact full gradients through the purification trajectory.

We further re-evaluate DiffAttack \citep{kang2024diffattack} by applying our standalone validation pipeline under the same CIFAR-10 setting, obtaining an adversarial accuracy of 61.57\%. In contrast, MEFA reduces adversarial accuracy to 42.75\%, an 18.82 percentage-point decrease over DiffAttack \citep{kang2024diffattack}. We also provide the first full-gradient evaluation of the EBM defense \citep{hill2021stochastic}; MEFA reduces its adversarial accuracy to 6.08\%, compared with 54.90\% under the original BPDA+EOT20 evaluation on CIFAR-10. On ImageNet, MEFA further reduces adversarial accuracy to 22\% and 34\% for ResNet-50 and WideResNet-50-2 classifiers purified by the score SDE VP-SDE defense under PGD+EOT20 $\ell_{\infty}$ attacks with $\varepsilon=4/255$.

% \vspace{-0.1in}
% \subsection{Experiment Settings}
% \vspace{-0.1in}
\textbf{Experimental Settings.} We will first present the settings for the numerical experiments.

% \vspace{-0.05in}
\textit{Dataset and Classifier.}
We randomly pick 510 images from CIFAR-10 \citep{krizhevsky2009learning} and consistently use them for all experiments, and 100 images from ImageNet \citep{deng2009imagenet} for the sake of fairness. 
The WideResNet-28-10 classifier \citep{DBLP:journals/corr/ZagoruykoK16} is used for the evaluation on CIFAR-10. ResNet-50 \citep{he2016deep} and Wideresnet-50-2 \citep{dosovitskiy2020image} are used for the evaluation on ImageNet. The original accuracy of these CIFAR-10 and ImageNet images is 95.10\% and 90\%, respectively.

\textit{Defense Evaluation Metric.}
To rigorously assess the robustness of adversarial defense mechanisms, we adopt the following standardized metrics, aligned with established practices in adversarial attacks: 
 % \citep{aleks2017deep, hill2021stochastic}
\begin{itemize}
\vspace{-0.12in}
    \item \textit{Original Accuracy \textbf{(OA)}}: Classification accuracy on the original unperturbed validation set.
    
\vspace{-0.05in}
    \item \textit{Natural Accuracy \textbf{(NA)}}: Classification accuracy of the original unperturbed and purified validation set with the defend model, also termed \textit{Clean Accuracy}.
    
\vspace{-0.05in}
    \item \textit{Adversarial Accuracy \textbf{(AA)}}: Classification accuracy under adversarial perturbations within threat model \(\mathcal{T}\). Also termed \textit{Robust Accuracy}. The final AA is re-evaluated using the saved adversarial images with the validation process.
\end{itemize}
\vspace{-0.1in}

\textit{Defense Model and Method.}
We use pre-trained diffusion models including discrete DDPM-based \citep{song2020denoising} and continuous score SDE-based \citep{song2020score}, non-smooth EBM models \citep{hill2021stochastic} to purify the images following \citet{blau2022threatmodelagnosticadversarialdefense, nie2022diffusion, hill2021stochastic} respectively. We trained the smooth EBM using the same model structure with softleakyrelu as the activation function. The reason for this activation choice is stated in  \Cref{sec:non_zero}.

\textit{Threat Model Specification.}
All metrics assume explicit adversarial constraints:
\begin{itemize}
\vspace{-0.12in}
    \item \textit{Perturbation Norm}: \(\ell_\infty\) or \(\ell_2\) on image scale of $[0,1]^D$.
    
\vspace{-0.05in}
    \item \textit{Attack Strength}:  \(\varepsilon = 8/255\) for \(\ell_\infty\), \(\varepsilon = 0.5 \) for \(\ell_2\) attacks on CIFAR-10.  \(\varepsilon = 4/255\) for \(\ell_\infty\) on ImageNet.
    
\vspace{-0.05in}
    \item \textit{Attack Method}: White-box Iterative Attacks. We use PGD \citep{aleks2017deep} with fixed step size of 2/255, and APGD from autoattack \citep{croce2020reliable} with adaptive step size on image scale of $[0,1]^D$. 
\end{itemize}
\vspace{-0.1in}

Prioritizing loss-optimized perturbations versus first- or final-state attack success instances introduces non-negligible variance. 20 attack replicates are used for all APGD, PGD and BPDA attacks, denoted as APGD+EOT20, PGD+EOT20, and BPDA+EOT20. APGD with loss-optimized and PGD with final adversarial state show very close results on $\ell_{\infty}$ as shown in \Cref{tab:table1,tab:table2} for Score SDE-based and EBM-based defenses. Thus, PGD is being used for our experiments other than \Cref{tab:table1}. We make sure the gradient checkpointing gradients match with the exact gradients close to $10^{-6}$--$10^{-7}$ on the defense model output without attacks. Iterative attacks with 100 steps including $\ell_{\infty}$ ($\varepsilon$ = 8/255) and $\ell_2$ ($\varepsilon$ = 0.5) are carried out on CIFAR-10. Iterative attacks with 50 steps including $\ell_{\infty}$ ($\varepsilon$ = 4/255) are carried out on ImageNet. More details of the experiment can be found in \Cref{sec:experiment}. The code is available online anonymously for review at \url{https://anonymous.4open.science/r/MEFA-24DF/}.

% \vspace{-0.1in}
% \subsection{Attack Against Continuous Diffusion Models}
% \vspace{-0.1in}

\vspace{-0.04in}\textbf{Attack Against Continuous Diffusion Models.} 
We use DiffPure \citep{nie2022diffusion} as the baseline score SDE-based defense because it reports SOTA robustness. Unlike DiffPure, which uses the adjoint method to estimate gradients under memory constraints, and DiffAttack \citep{kang2024diffattack}, which combines adjoint-based gradients with a deviated-reconstruction loss, MEFA evaluates the defense with exact full gradients. As shown in \Cref{tab:table1,tab:table2}, MEFA achieves SOTA attack performance against the continuous diffusion, score SDE-based defense \citep{nie2022diffusion}, reducing adversarial accuracy to 42.75\% and 41.76\% under PGD and APGD $\ell_{\infty}$ attacks with $\varepsilon=8/255$, respectively. The loss-optimized and final adversarial states yield similar results, while BPDA+EOT20 fails to effectively attack the score SDE-based defense, as shown in \Cref{tab:table3}. On ImageNet, MEFA also substantially reduces adversarial accuracy compared with DiffPure \citep{nie2022diffusion}, as reported in \Cref{tab:imagenet_diffpure}. These results across CIFAR-10 and ImageNet highlight the importance of exact-gradient evaluation over approximate-gradient baselines for reliable robustness assessment.

% \vspace{-0.1in}
% \subsection{Attack Against Discrete Diffusion Models}
% \vspace{-0.1in}
\vspace{-0.04in}\textbf{Attack Against Discrete Diffusion Models.} 
Similarly, our evaluation shows that DDPM-based defense has a weaker robustness compared to the score SDE-based defense, with adversarial accuracy of 35.48\% and 68.23\% against $\ell_{\infty}$ and $\ell_2$ PGD attacks in \Cref{tab:table2}. This is significantly lower than Diffattack's evaluation method of 67.65\% on APGD $\ell_\infty$ attacks. SOTA accuracy of 35.29\% is shown for APGD $\ell_\infty$ attacks as well in \Cref{tab:table1}. 
% Similar adversarial accuracy of 34.91\% is also shown in \Cref{tab:table7} on WRN-70-16. 
Similarly, BPDA + EOT20 in \Cref{tab:table3} failed to attack against DDPM-based defenses.

\begin{table*}[t]
\centering
\begin{minipage}[t]{0.61\textwidth}
\vspace{0pt}
\centering
\caption{\small Defense against PGD+EOT20 attacks from WRN-28-10 on CIFAR-10.}
\label{tab:table2}
\vspace{0.02in}
\scriptsize
\renewcommand{\arraystretch}{1.15}
\setlength{\tabcolsep}{2.2pt}
\resizebox{\linewidth}{!}{
\begin{tabular}{p{3.5cm}ccccc}
\toprule
\textbf{Defense} & \textbf{Attack} & \multicolumn{2}{c}{\textbf{Nat. Accuracy (NA)}} & \multicolumn{2}{c}{\textbf{Adv. Accuracy (AA)}} \\
\cmidrule(lr){3-4} \cmidrule(lr){5-6}
\textbf{Model} & \textbf{Type~~\,} & $H_{\mathrm{d}}=1$ & $H_{\mathrm{d}}=50$ & $H_{\mathrm{d}}=1$ & $H_{\mathrm{d}}=50$ \\
\midrule
\multirow{2}{3.5cm}{Score SDE VP-SDE \\(Contin. Diffusion)} 
& $\ell_\infty$ & 88.82\% & 93.33\% & 43.53\% & \bftab{41.76\%} \\
& $\ell_2~\,$ & 90.79\% & 94.62\% & 70.00\% & 77.65\% \\
\midrule
\multirow{2}{3.5cm}{2DUnet DDPM \\(Discrete Diffusion)} 
& $\ell_\infty$ & 82.55\% & 90.79\% & 41.56\% & \bftab{35.48\%} \\
& $\ell_2~\,$ & 82.74\% & 89.81\% & 64.71\% & 68.23\% \\
\midrule
\multirow{2}{3.5cm}{Smooth EBM} 
& $\ell_\infty$ & 75.88\% & 84.51\% & ~\,8.43\% & ~\,\bftab{6.08\%} \\
& $\ell_2~\,$ & 75.88\% & 84.51\% & 41.96\% & 41.76\% \\
\bottomrule
\end{tabular}
}
\end{minipage}
\hfill
\begin{minipage}[t]{0.37\textwidth}
\vspace{0pt}
\centering
\caption{\small Score SDE VP-SDE defense against PGD+EOT20 $\ell_\infty$ attacks ($\epsilon=4/255$) on ImageNet.}
\label{tab:imagenet_diffpure}
\vspace{0.02in}
\scriptsize
\renewcommand{\arraystretch}{1.15}
\setlength{\tabcolsep}{3pt}
\resizebox{\linewidth}{!}{
\begin{tabular}{llcc}
\toprule
\textbf{Threat Model} & \textbf{Attack Method} & \textbf{Nat. Acc} & \textbf{Adv. Acc} \\
\midrule
\multirow{3}{*}{ResNet-50} 
& Ours, $H_d{=}1$  & 75.00\% & 24.00\% \\
& Ours, $H_d{=}50$ & 77.00\% & \bftab{22.00\%} \\
\cmidrule(lr){2-4}
& DiffPure* \citep{nie2022diffusion} & 67.79\% & 40.93\% \\
\midrule
\multirow{3}{*}{WideResNet-50-2} 
& Ours, $H_d{=}1$  & 82.00\% & \bftab{33.00\%} \\
& Ours, $H_d{=}50$ & 81.00\% & 34.00\% \\
\cmidrule(lr){2-4}
& DiffPure* \citep{nie2022diffusion} & 71.16\% & 44.39\% \\
\bottomrule
\end{tabular}
}

\vspace{0.5mm}
\scriptsize
\emph{Note:} * reported from the original paper.
\end{minipage}

\vspace{-2mm}
\end{table*}

\vspace{-0.05in}
\textbf{Attack Against EBM-Based Defenses.} 
We use the long-run EBM defense with 1500 Langevin steps \citep{hill2021stochastic} as the EBM baseline because it reports SOTA robustness among EBM-based defenses, as shown in \Cref{tab:table1,tab:table2}. Prior evaluations rely on BPDA+EOT20 due to the memory cost of differentiating through long EBM sampling; in our re-evaluation, BPDA+EOT20 yields 49.8\% adversarial accuracy on the smooth-activation EBM (\Cref{tab:table3}), compared with 54.9\% reported by Stochastic Security \citep{hill2021stochastic} using 150 defense replicates. MEFA provides the first full-gradient attack evaluation of this EBM defense. The original non-smooth pretrained EBM is re-evaluated with comparable robust accuracy of 53.5\% using 50 replicates; additional details are provided in \Cref{sec:EBM_comparison}.

\begin{figure}
    \centering
    \includegraphics[width=.8\textwidth]{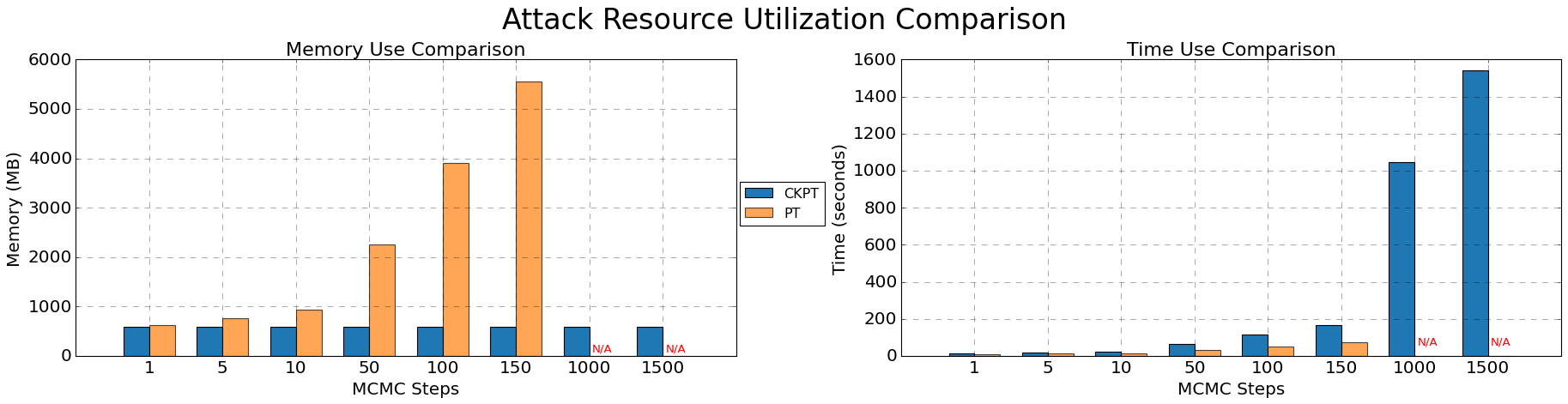}
    \vspace{-0.12in}
    \caption{\small Memory(left), time(right) comparison between MEFA Framework's gradient checkpointing (CKPT) and standard Pytorch (PT) backpropagation for smooth EBM against PGD $\ell_\infty$ attack on one CIFAR-10 image with WideResNet-28-10.}
    \label{fig:memry_time}
    \vspace{-0.15in}
\end{figure}

% \vspace{-0.1in}
% \subsection{Memory and Time Comparison: MEFA-PGD vs. Standard PGD}
\vspace{-0.05in}
\textbf{Memory and Time Comparison: MEFA Framework-PGD vs. Standard PGD.} 
We validate the memory cost using a smooth EBM defense model to fit the fixed memory compared to the standard PGD attack. \Cref{fig:memry_time} shows that our MEFA-PGD maintains a constant memory use cost and doubles the time usage on the same MCMC step since the backpropagation recomputes the intermediate states. In contrast, the standard PGD backpropagation, which had the intermediate image states on the graph, increases the memory use linearly and is unable to fit in the GPU memory after 150 steps. Thus, no bars are shown for the standard PDG attack at the 1000th and 1500th step. 
% The Google Colab T4 GPU with 16GB of memory is used for the demonstration.  

\begin{table*}[t]
\centering
\scriptsize
\setlength{\tabcolsep}{2.5pt}
\renewcommand{\arraystretch}{1.15}

\begin{minipage}[t]{0.56\textwidth}
\centering
\caption{\small Defense against BPDA+EOT20 attacks on CIFAR-10}
\label{tab:table3}

\resizebox{\linewidth}{!}{
\begin{tabular}{p{3cm}ccccc}
\toprule
\textbf{Defense} & \textbf{Attack} & \multicolumn{2}{c}{\textbf{Nat. Accuracy (NA)}} & \multicolumn{2}{c}{\textbf{Adv. Accuracy (AA)}} \\
\cmidrule(lr){3-4} \cmidrule(lr){5-6}
\textbf{Model} & \textbf{Type~~\,} & \textbf{$H_d=1$} & \textbf{$H_d=50$} & \textbf{$H_d=1$} & \textbf{$H_d=50$} \\
\midrule
\multirow{2}{5cm}{Score SDE VP-SDE \\ (Contin. Diffusion)}
& $\ell_\infty$ & 89.22\% & 93.72\% & 80.39\% & 86.08\% \\
& $\ell_2~\,$ & 88.82\% & 93.33\% & 83.33\% & 89.80\% \\
\midrule
\multirow{2}{5cm}{2DUnet DDPM \\ (Discrete Diffusion)}
& $\ell_\infty$ & 81.96\% & 90.78\% & 70.98\% & 76.47\% \\
& $\ell_2~\,$ & 82.94\% & 90.20\% & 78.63\% & 85.10\% \\
\midrule
\multirow{2}{5cm}{Smooth EBM}
& $\ell_\infty$ & 75.88\% & 84.51\% & 44.14\% & 49.80\% \\
& $\ell_2~\,$ & 75.88\% & 84.51\% & 63.33\% & 73.13\% \\
\bottomrule
\end{tabular}
}
\end{minipage}
\hfill
\begin{minipage}[t]{0.42\textwidth}
\centering
\caption{\small OOD Defense against PGD+EOT20 and BPDA+EOT20 $\ell_\infty$ attacks on CIFAR-10}
\label{tab:table5}

\resizebox{\linewidth}{!}{
\begin{tabular}{llcccc}
\toprule
\textbf{Attack} & \textbf{Dataset} & \multicolumn{2}{c}{\textbf{Nat. Accuracy (NA)}} & \multicolumn{2}{c}{\textbf{Adv. Accuracy (AA)}} \\
\cmidrule(lr){3-4} \cmidrule(lr){5-6}
\textbf{Method} & \textbf{Pre-trained} & \textbf{$H_d=1$} & \textbf{$H_d=50$} & \textbf{$H_d=1$} & \textbf{$H_d=50$} \\
\midrule
PGD+EOT20 & \multirow{2}{*}{FOOD} & \multirow{2}{*}{71.37\%} & \multirow{2}{*}{82.75\%} & 13.92\% & ~\,7.65\% \\
\cmidrule(lr){5-6}
BPDA+EOT20 & & & & 38.63\% & 36.86\% \\
\midrule
PGD+EOT20 & \multirow{2}{*}{CINIC10} & \multirow{2}{*}{83.01\%} & \multirow{2}{*}{90.43\%} & 41.41\% & 37.89\% \\
\cmidrule(lr){5-6}
BPDA+EOT20 & & & & 47.45\% & 43.33\% \\
\bottomrule
\end{tabular}
}
\end{minipage}

\vspace{-0.08in}
\end{table*}

% \subsection{OOD Insights on Defense Robustness}\label{sec:action}

\textbf{OOD Insights on Defense Robustness.} 
We examine the OOD defense accuracy for DDPM-based defenses since it's shown better defense outcomes than EBM and it's faster to train than score SDE-based diffusion models. The OOD accuracy depends on the pre-trained data distribution. Thus, a defense model with diverse datasets could have the potential to secure the classifier. It shows that the defense accuracy is highly dependent on the data distribution on which the model is trained. The defense model trained on CINIC10, which is a combination of CIFAR-10 and ImageNet, has a comparable adversarial accuracy (37.89\% vs. 35.48\%) with CIFAR-10 trained defense model, and a better adversarial accuracy compared to the defense model trained on FOOD shown in \Cref{tab:table5}.

\vspace{-0.1in}
\section{Related Work}
\vspace{-0.15in}

\textbf{Adversarial Attack.}
Adversarial attacks expose the vulnerability of modern models \citep{szegedy2013intriguing} across white-box \citep{IanGan2014,aleks2017deep,carlini2017evaluatingrobustnessneuralnetworks,croce2020reliable,croce2021mind}, black-box \citep{papernot2017practicalblackboxattacksmachine,brendel2018decisionbasedadversarialattacksreliable,Su_2019,cai2021zobcd,kim2025cars}, and physical settings \citep{eykholt2018robustphysicalworldattacksdeep}, with adaptive white-box attacks serving as the strongest empirical test. Recent works further strengthen adaptive attacks against diffusion-based purification defenses through improved gradient optimization and memory-efficient techniques such as checkpointing \citep{kassis2024diffbreak,li2024adbm,liu2024towards}. However, robustly evaluating iterative stochastic purification defenses \citep{kang2024diffattack,lee2023robust,chen2023robust}, including diffusion- and EBM-based methods, remains challenging due to memory constraints and stochastic instability. MEFA addresses these limitations by enabling exact-gradient evaluation with controlled stochastic validation.

\vspace{-0.001in}\textbf{Adversarial Purification (AP).} Early AP methods employ generative models such as GANs \citep{samangouei2018defense} or VAEs \citep{yoon2021adversarial} to project corrupted inputs onto the clean data manifold. Recent advances leverage stochastic purification mechanisms, including randomized smoothing \citep{cohen2019certified}, diffusion models \citep{nie2022diffusion}, energy-based models \citep{grathwohl2019modeling,du2019implicit,hill2021stochastic}, and score-based Langevin dynamics \citep{yoon2021adversarial}. While these methods improve robustness, they are particularly susceptible to adaptive attacks when gradients are accurately computed \citep{athalye2018obfuscated}. Our work complements this line of research by providing a more faithful evaluation framework for such stochastic defenses, revealing their true robustness under strong white-box attacks.

\vspace{-0.1in}
\section{Conclusion}
\vspace{-0.15in}
Our work establishes the Memory-Efficient Full-gradient Attack (MEFA) framework as a rigorous benchmark for evaluating iterative stochastic purification defenses, with a focus on diffusion- and EBM-based methods. By enabling full-gradient computation with $\mathcal{O}(1)$ memory cost, MEFA reveals substantial vulnerabilities in both defense families and highlights the need for standardized evaluation of iterative stochastic purification. Overall, MEFA helps bridge the gap between theoretical robustness claims and practical deployability by quantifying the trade-offs among memory constraints, computational overhead, and stochastic outcome variance. 
\medskip

\bibliographystyle{plainnat}
\renewcommand{\bibsection}{\subsubsection*{References}}
\bibliography{references}

%%%%%%%%%%%%%%%%%%%%%%%%%%%%%%%%%%%%%%%%%%%%%%%%%%%%%%%%%%%%
\newpage 

\appendix

%%%%%%%%%%%%%%%%%%%%%%%%%%%%%%%%%%%%%%%%%%%%%%%%%%%%%%%%%%%%
\begin{center}
\LARGE Supplementary Materials for \\
\textbf{Memory Efficient Full-gradient Attacks (MEFA) Framework for Adversarial Defense Evaluations}
\end{center}

\section{MEFA Framework - PGD} \label[appendix]{sec:FullGradAttack}

We summarize the full gradient attack algorithm for MEFA Framework PGD with fixed step size used for adversarial attacks in \Cref{alg:FullGradAttack}. The APGD attack with adaptive step size is based on the framework of autoattack \citep{croce2021mind} with the same hyperparameters. However, the gradient is on the graph during the backpropagation for a more precise calculation.

\begin{algorithm}
\caption{MEFA-PGD Attack Algorithm}\label{alg:FullGradAttack}
\small{
\begin{algorithmic}[1]
%\hspace*{\algorithmicindent} 
\State \textbf{Input:} {Natural images $\{\mathbf{x}^+_m\}_{m=1}^{M}$, defense model $T(\mathbf{x};\theta)$, classifier  $f$,  $l_{\text{norm}}$ radius $\varepsilon$, attack step size $\eta$}
%\hspace*{\algorithmicindent} 
\State \textbf{Output:} {$\tilde{\mathbf{x}}$}

\For{$1 \leq i \leq M$}
\State{$\text{select} (\mathbf{x}_i, y_i) \text{from batch}$}
\State{Randomly initialize adversary $\tilde{\mathbf{x}}^{(0)}$ inside $\ell_{\text{norm}}$ ball around $\mathbf{x}_i$}
\State{$\tilde{\mathbf{x}}^{(0)} \gets \mathbf{x}_i$}

\For{$1 \leq j \leq N$}
\State{$ \mathcal{L}(\tilde{\mathbf{x}}^{(j)}, y_i) \gets F(\tilde{\mathbf{x}}^{(j)}) = \mathbb{E}_{T(\tilde{x}^{(j)})} [f(T(\tilde{\mathbf{x}}^{(j)}))]$} \Comment{calculate loss and logits with \eqref{eqn:predict_label}}
\State{$\nabla_{\tilde{\mathbf{x}}^{(j)}} = \nabla_\text{CKPT} (\tilde{\mathbf{x}}^{(j)}, y_i )$} \Comment{calculate attack gradient with \Cref{alg:GradientCheckpointing}}
\State{$\tilde{\mathbf{x}}^{(j+1)} \gets \text{PGD} (\tilde{\mathbf{x}}^{(j)} + \eta \nabla_{\tilde{\mathbf{x}}^{(j)}}(\mathcal{L}(\tilde{\mathbf{x}}^{(j)}, y_i) )$}  \Comment{update the adversarial state with \eqref{eqn:PGD}}

\EndFor
\EndFor
\end{algorithmic}
}
\end{algorithm}

\section{Gradient Checkpointing Algorithm for Adversarial Attack} \label[appendix]{sec:GradientCheckpointing}

We summarize the gradient checkpointing Algorithm for Adversarial Defense in \Cref{alg:GradientCheckpointing}:

\begin{algorithm}  
\caption{Gradient Checkpointing Algorithm (PyTorch-like Pseudo-codes)}\label{alg:GradientCheckpointing}  
\begin{algorithmic}[1]  
%\hspace*{\algorithmicindent} 
\State \textbf{Input:} defense model $D$, $\mathbf{x}_i$ ($i \in [T]$), $\frac{\partial \mathcal{L}}{\partial \mathbf{x}'_0}$ , T:number of sampling steps, noise $\mathbf{z_i} \sim \mathcal{N}(0,I)$ 
%\hspace*{\algorithmicindent} 
\State \textbf{Output:} $\frac{\partial \mathcal{L}}{\partial \mathbf{x}_0}$  

\State  {\text{Initialize} $hs \gets \text{list of zeros to store intermediate images } $}
\State  {\text{Initialize} $zs \gets \text{list of zeros to store intermediate noises } $}
    \For{$t = T \text{ to } 0$}  \Comment{forward pass for reverse sampling}  
        \State {\text{Detach\_Device}($D(\mathbf{x}_t,\mathbf{z}_t) \to \mathbf{x}'_{t-1}$)  }
    \EndFor
    
    \For{$k \in [T]$}  \Comment{backpropagation}
        \State{$\mathbf{x}_t = hs[n-k-1], \mathbf{z}_t = zs[n-k-1]$}
        \State {\text{Create\_GraphOnDevice}($D(\mathbf{x}_t,\mathbf{z}_t) \to \mathbf{x}_{t-1}$)  }
        \State {$\frac{\partial \mathcal{L}}{\partial \mathbf{x}_t} \gets \text{auto\_grad}(\mathbf{x}_{t-1}, \mathbf{x}_t, \frac{\partial \mathcal{L}}{\partial \mathbf{x}'_0})$ }
        \State {$\frac{\partial \mathcal{L}}{\partial \mathbf{x}'_0} \gets \frac{\partial L}{\partial \mathbf{x}_t}$ }  
    \EndFor
\end{algorithmic}  
\end{algorithm}  

\section{Comparison Between Gradient Checkpointing and Segment-Wise Forwarding-Backwarding Algorithms}\label[appendix]{sec:algorithm_comparison}
Strong adaptive attacks require computing the full gradients of the defense system. Gradient computation in deep neural networks, particularly for iterative generative models such as EBM and diffusion models, is often memory intensive due to the need to retain intermediate results on the graph for backpropagation. The segment-wise forward-backward algorithm from the DiffAttack Method \citep{kang2024diffattack} and our gradient checkpointing algorithm are based on the similar mechanisms. In the following, we provide a detailed comparison of the two methods.

Both gradient checkpointing \Cref{alg:GradientCheckpointing} and segment-wise forwarding-backwarding algorithm \citep{kang2024diffattack} are designed to balance memory efficiency and computational overhead during gradient computation for generative model sampling process. While both recompute forward steps during backpropagation, they differ in their memory management strategies and the granularity of operations.

Gradient checkpointing \Cref{alg:GradientCheckpointing} minimizes memory usage by detaching intermediate states during the forward pass. These detached states are later recomputed during the backward pass to create the necessary computational graph for gradient computation. This approach trades memory efficiency for increased computational overhead, as every intermediate state must be recomputed.

Segment-wise forwarding-backwarding operates by breaking the computation into segments. During the backward pass, forward states are recomputed for each segment using the saved intermediate results. While this also involves recomputation, it limits the size of the computational graph during backpropagation by processing each segment independently, which can improve memory efficiency. It avoids explicitly applying the chain rule step-by-step through each noise level or intermediate state. Instead, it leverages recomputed gradients and deviated-reconstruction loss (mse+ce) at multiple layers as the surrogate loss to approximate the effect of the chain rule. It does not represent the true gradient in a strict mathematical sense.

Segment-wise forwarding-backwarding \citep{kang2024diffattack} is not a full gradient attack, since the final gradients are approximated using deviated-reconstruction loss (mse+ce) making this a projection rather than a full backward pass for the DDPM-based defenses. For score SDE-based defenses, gradient is computed with single backward pass from a combined reconstructed loss (mse+ce) with all intermediate states and the adjoint method is used. The adjoint method uses scalable gradients with memory-efficient noise caching \citep{li2020scalable}. However, it is an approximation of the gradients and doesn't allow the attack to backpropgate through the entire sampling chain. Our method directly uses the SDE solver termed sdeint \citep{kloeden1992stochastic} by gradient checkpointing to compute the gradients for PGD attack. See the specific code application comparison in \Cref{tab:grad_comparison} and \Cref{tab:ddpm_grads}.
 % ===== CODE STYLE DEFINITIONS =====
\definecolor{codegreen}{rgb}{0,0.6,0}
\definecolor{codegray}{rgb}{0.5,0.5,0.5}
\definecolor{codepurple}{rgb}{0.58,0,0.82}
\definecolor{backcolour}{rgb}{0.95,0.95,0.92}
\definecolor{codealert}{rgb}{1,0,0} % Added red color definition
% DiffAttack Style
 \lstdefinestyle{diffattack}{
    backgroundcolor=\color{backcolour},
    commentstyle=\color{codegreen},
    keywordstyle=\color{magenta},
    numberstyle=\tiny\color{codegray},
    stringstyle=\color{codepurple},
    basicstyle=\ttfamily\footnotesize,
    breakatwhitespace=false,         
    breaklines=true,
    frame=single,
    numbers=left,
    emph={torch.autograd.grad}, % Added emphasis rule
    emphstyle=\color{codealert}, % Red color for emphasized items
    moredelim=[is][\color{codealert}]{@**@}{@**} % Red line highlighting
}

% MEFA Style 
\lstdefinestyle{MEFA}{
    backgroundcolor=\color{backcolour},
    commentstyle=\color{codeblue},
    keywordstyle=\color{red},
    numberstyle=\tiny\color{codegray}, % Add this line
    basicstyle=\ttfamily\footnotesize,
    breaklines=true,
    % frame=tb,
    frame=single,
    numbers=right,
    emph={torch.autograd.grad}, % Added emphasis rule
    emphstyle=\color{codealert}, % Red color for emphasized items
    moredelim=[is][\color{codealert}]{@**@}{@**} % Red line highlighting
}

 \begin{table}[ht]
\centering
\caption{Score SDE-based Gradient Calculation: DiffAttack vs. MEFA Framework}
\label{tab:grad_comparison}
\begin{tabular}{p{0.48\textwidth}p{0.48\textwidth}}
\toprule
\textbf{DiffAttack (Segment-wise Gradient)} & \textbf{MEFA (Full Gradient Checkpointing)} \\
\midrule
\begin{lstlisting}[style=diffattack]
# DiffAttack Code: Approximate gradients with deviated-reconstruction loss
logits, mid_x, ori_x = self.model(
    x_adv, 
    return_mid=True
)
loss_indiv = self.mse_loss(
    mid_x, 
    ori_x, 
    target_ind
)
if len(loss_indiv.shape) > 3:
    loss_indiv = loss_indiv.view(
        loss_indiv.shape[0], 
        loss_indiv.shape[1], -1
    )
loss_indiv = torch.mean(loss_indiv, dim=-1)
loss_indiv = torch.mean(loss_indiv, dim=0)
loss_indiv = loss_indiv.view(x.shape[0])
loss_ce = criterion_indiv(logits, y)
loss_indiv += loss_ce
loss = torch.mean(loss_indiv)
@**@grad = torch.autograd.grad(
    loss, 
    [x_adv]
)[0].detach()@**
\end{lstlisting}
&
\begin{lstlisting}[style=MEFA]
# ME framwork Code: Full gradient computation, input: CE loss
net_grads = torch.autograd.grad(
    loss, 
    [X_repeat_purified]
)[0]
curr_t = curr_ts[n-i-1].to(device)
next_t = next_ts[n-i-1].to(device)
net_out = ys[n-i-1]
curr_y = torch.autograd.Variable(
    net_out, 
    requires_grad=True
).to(device)
noi = noises[n-i-1].to(device)
prev_t, prev_y = curr_t, curr_y
next_layer, curr_extra = self.step(
    curr_t, 
    next_t, 
    curr_y, 
    noi, 
    curr_extra
)
curr_t = next_t
@**@net_grads = torch.autograd.grad(
    next_layer, 
    [curr_y], 
    grad_outputs=net_grads
)[0]@**
grad = net_grads.view(
    X_repeat_purified.shape
)
\end{lstlisting} \\
\bottomrule
\end{tabular}

\vspace{2mm}
\footnotesize\textbf{Key Technical Differences for Score SDE-based Defense:}
\begin{itemize}
    \item Left: Approximates gradients through segmented forward/backward passes via adjoint method with deviated-reconstruction
    \item Right: Computes exact gradients from CE loss via SDE solver gradient checkpointing with $\mathcal{O}(1)$ memory
    \item \textcolor{codealert}{Red lines}: Core gradient computation steps
\end{itemize}
\end{table}

For score SDE-based defenses, an augmented SDE for calculating the gradient of an objective function $L$ w.r.t. the input $\mathbf{x'_T}$ of the SDE is given by
\begin{align}
\frac{\partial \mathcal{L}}{\partial \mathbf{x'_T}} &= \text{sdeint} \left(
    \frac{\partial \mathcal{L}}{\partial {\mathbf{x'_0}}}, 
    {\mathbf{x'_0}}, 
    \mathbf{f}, 
    \mathbf{g}, 
    \mathbf{w}, 
    \mathbf{0}, 
    T
\right),
\end{align}
where $\frac{\partial \mathcal{L}}{\partial {\mathbf{x'_0}}}$
is the gradient of the objective \(\mathcal{L}\) with respect to the output \(\mathbf{x'_0}\) of the SDE in \Cref{eqn:ldiff}, and 
\[
\mathbf{f}([\mathbf{x}; \mathbf{z}], t) =
\begin{pmatrix}
    f_{\text{rev}}(\mathbf{x}, t) \\
    \frac{\partial f_{\text{rev}}(\mathbf{x}, t)}{\partial \mathbf{x}} \mathbf{z}
\end{pmatrix},
\]
\[
\mathbf{g}(\mathbf{t}) =
\begin{pmatrix}
    -g_{\text{rev}}(t) \mathbf{1}_d \\
    \mathbf{0}_d
\end{pmatrix},
\]
\[
\mathbf{w}(t) =
\begin{pmatrix}
    -\mathbf{w}(1 - t) \\
    -\mathbf{w}(1 - t)
\end{pmatrix},
\]
with \(\mathbf{1}_d\) and \(\mathbf{0}_d\) representing the \(d\)-dimensional vectors of all ones and all zeros, respectively.

\begin{table}[ht]
\centering
\caption{DDPM-based Gradient Calculation: DiffAttack vs. MEFA Framework}
\label{tab:ddpm_grads}
\begin{tabular}{p{0.48\textwidth}p{0.48\textwidth}}
\toprule
\textbf{DiffAttack (Segment-wise Gradient)} & \textbf{MEFA (Full Gradient Checkpointing)} \\
\midrule
\begin{lstlisting}[style=diffattack]
# DiffAttack Code: Gradient approximation, input: grad
if self.total_noise_levels >= 1:
    # Forward diffusion process
    out = ori_x * self.a[self.total_noise_levels - 1].sqrt() 
        + self.e * (1.0 - self.a[ self.total_noise_levels - 1]).sqrt()
    
# Gradient computation with weighted loss
loss = torch.sum(out * grad)
@**@grad_new = torch.autograd.grad(
        loss, 
        [ori_x]
    )[0].detach()@**
\end{lstlisting}
&
\begin{lstlisting}[style=MEFA]
# MEFA Framework Code: Full gradient computation, input: net_grads
net_out = x_diff_list[n - k - 1]
net_out = torch.autograd.Variable(
    net_out, 
    requires_grad=True
).to(model.device)
noi_out = noi_diff_list[
    n - k - 1
].to(model.device)
t = torch.tensor(
    [n - k - 1] * shape[0], 
    device=X.device
)

# DDPM sampling forward process
next_layer = scheduler.ddim_sample_grad(
    model,
    net_out,
    t,
    noi_out,
    model_kwargs=model_kwargs,
    eta=args.eta
)
next_layer_out = next_layer["sample"]
@**@net_grads = torch.autograd.grad(
    next_layer_out,
    [net_out],
    grad_outputs=net_grads
)[0]@**
\end{lstlisting} \\
\bottomrule
\end{tabular}

\vspace{2mm}
\footnotesize\textbf{Key Technical Differences for DDPM-based Defense:}
\begin{itemize}
    \item Left: Approximates gradients at specific noise levels using deviated-reconstruction loss.
    \item Right: Computes exact gradients from CE loss through full DDPM sampling chain with $\mathcal{O}(1)$ memory.
    \item \textcolor{codealert}{Red lines}: Core gradient computation steps.
\end{itemize}
\end{table}

\noindent
\section{Experiment Details}\label[appendix]{sec:experiment}
In our experiments, we use pre-trained score SDE-based diffusion model \citep{song2020score} on CIFAR-10 and ImageNet, DDPM-based diffusion model \citep{song2020denoising} and EBM-based model \citep{hill2021stochastic} on CIFAR-10 according to the defense methods from \citep{nie2022diffusion,blau2022threatmodelagnosticadversarialdefense,hill2021stochastic} for image purification. Both diffusion-based defenses use 100 denoising steps. EBM-based defenses use 1500 sampling steps. Given the significant computational requirements, we randomly select a fixed subset of 510 images from the validation set to ensure fair comparisons across all experiments in robust evaluations to avoid data selection randomness on CIFAR-10 and 100 images on ImageNet. The proposed MEFA Framework PGD attack is implemented within the PGD \citep{aleks2017deep} or AGPD framework \citep{croce2020reliable}, maintaining identical hyperparameters. Specifically, for both PGD and APGD, the number of attack iterations is set to 100 on CIFAR-10, and 50 on ImageNet, while the attack replicates for the gradient approximation EOT are fixed at 20. For PGD, the step size $\alpha$ is set to $2/255$. For the APGD, the momentum coefficient $\alpha$ is set to 0.75, and the step size $\eta$ is initialized as $2\varepsilon$, where $\varepsilon$ represents the maximum $\ell$-norm of the perturbations. For adversarial attacks, we evaluate with $\varepsilon = 8/255$ for $\ell_\infty$-norm, and $\varepsilon = 0.5$ for $\ell_2$-norm. 

For strong attacks such as $\ell_{\infty}$ ($\varepsilon$ = 8/255 on CIFAR-10) in order to compare with existing results, we use PGD with Adaptive adversarial steps (APGD) according to autoattack \citep{croce2020reliable}. For weaker attacks such as $\ell_2$ ($\varepsilon$ = 0.5), we use PGD to explore stable adversarial examples. We save the first broken state, final and/or loss-optimized final adversarial state during the attack process so that the validation process could validate the outcomes. All results in the tables are validated using the validation process. We chose the final adversarial state for PGD attacks and the loss-optimized final misclassified adversarial state for APGD from autoattack \citep{croce2020reliable} for the validation. 

We run our experiments on 4 workstations, each with 3 NVIDIA RTX A2000 GPUS with 12GB memory on CIFAR-10. EBM-based defense takes 4.5 days for ME PGD per experiment, while score SDE-based defense takes 9 days per experiment. We run our experiments on available Tesla V100 GPUS with 32GB memory on ImageNet. Google Colab T4 GPU with 16GB of memory is used for demonstrating the memory and time cost.

\subsection*{Attack  Steps}
\begin{figure}[ht]
\centering
    \includegraphics[width=.65\textwidth, height=5cm]{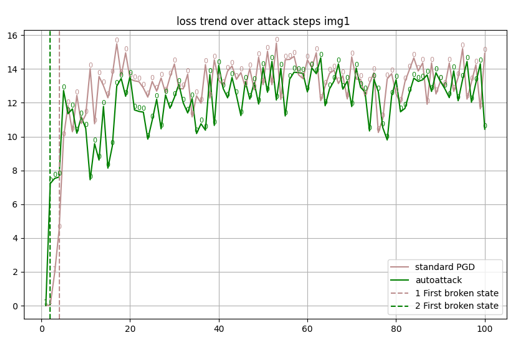}
    \caption{\small Loss trend stabilized at 20 attack steps. \textit{red line}: PGD loss \citep{aleks2017deep}, \textit{green line}: APGD loss from autoattack \citep{croce2020reliable}. \textit{0's on the lines}: wrong prediction. \textit{vertical dash lines}: first broken state}
    \label{fig:attack_loss}
\end{figure}

\label[appendix]{sec:attack_steps}
The loss trajectory of one image example stabilizes after 20 attack steps, and fluctuations could be handled during the validation process (see \Cref{fig:attack_loss}). The plot shows that the loss of the first broken states is much lower than the later stable states, thus inducing more volatility of the outcome evaluation.

\subsection*{Defense Replicates $H_\mathrm{d}$ Statistical Analysis}\label{sec:hdef_estimate}

\begin{figure}[ht]
\centering
    \includegraphics[width=.32 \textwidth]{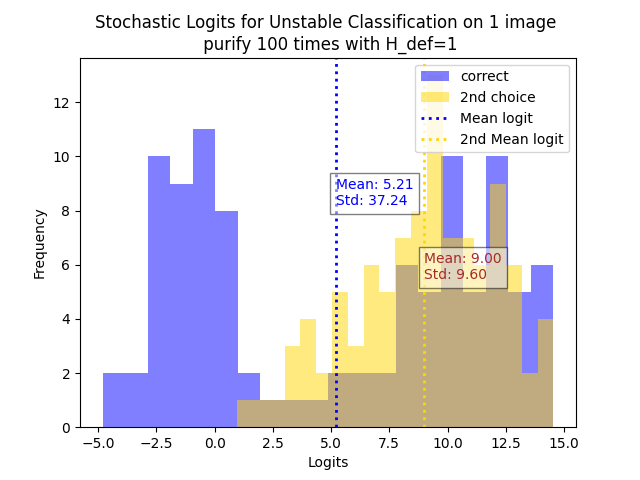}
    \includegraphics[width=.32\textwidth]{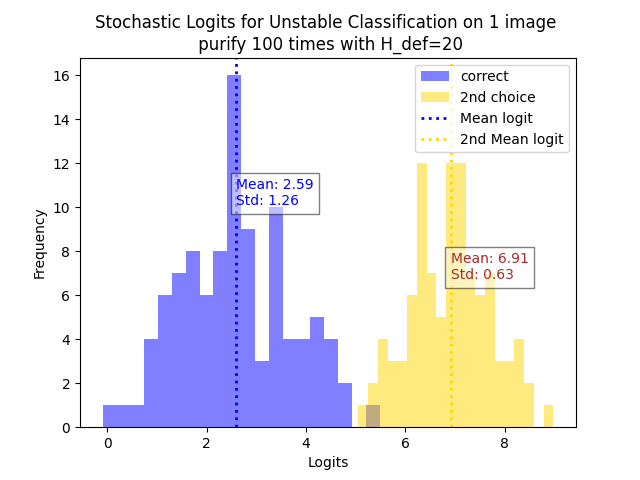}
    \includegraphics[width=.32\textwidth]{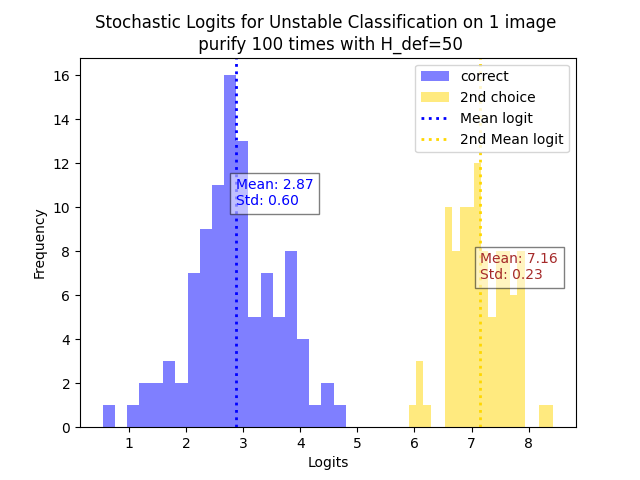}
    \vspace{-0.15in}
    \caption{\small Defense replicates required analysis to stabilize prediction due to purification randomness. Histogram of correct logits and highest incorrect logits, with mean and standard deviation for the adversarial image with different $H_\mathrm{d}$. \textit{Left}: $H_\mathrm{d}=1$.
    \textit{Middle}: $H_\mathrm{d}=20$.
    \textit{Right}: $H_\mathrm{d}=50$.}
    \label{fig:defense_replicates}
    \vspace{-0.1in}
\end{figure}

\Cref{fig:defense_replicates} shows that 50 replicates are sufficient when the correct logit is well separated from the largest incorrect logit across repeated purification trials.

\begin{figure}[ht]
\centering
    \includegraphics[width=.65\textwidth]{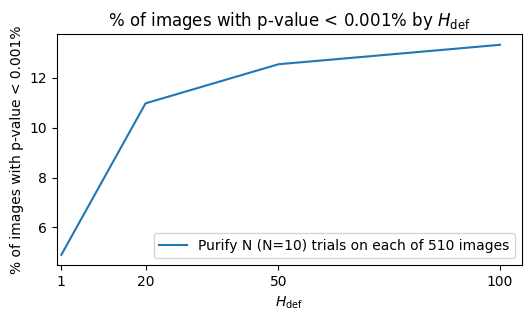}
    \caption{\small \% of images with p-value<0.001\% by $H_\mathrm{def} = 1, 20, 50, 100$}
    \label{fig:hdef_estimate}
\end{figure}

Additionally, we design the experiment by purifying each adversarial image $N$ trials using $H_\mathrm{d}$ of 1, 20, 50, 100. Then, the p-value is calculated for the distributions of correct logits and 2nd choice logits (the highest incorrect logits) from the classifier. The percentage of images whose p-value < 0.0001 is plotted in \Cref{fig:hdef_estimate}. $N$ is set to 10 in this plot. Larger $H_\mathrm{d}$ have a higher percentage of images whose p-value < 0.0001, which indicates a more stable prediction after the purification.

\subsection*{Non-zero Second Derivatives}\label[appendix]{sec:non_zero}
In the context of adversarial defense evaluation, we implement a rigorous approach to ensure that the gradient-checkpointing second-order derivatives align with those computed using the exact full gradients, within the constraints imposed by available memory. Specifically, we utilize smooth activation functions in our EBM, referred to as smooth EBM, which facilitates stable and accurate gradient computations. This choice is critical as it allows us to fairly evaluate the defense against the strong PGD attacks. The diffusion model utilizes the smooth activation function of SiLU. In the case of non-smooth activation functions such as Leaky ReLU, PGD and BPDA attacks will be equivalent due to the zero second-order derivative. 

\section{Additional comparison with other gradient approximation methods}\label{sec:other_methods}
We have conducted the experiments with Diffpure Adjoint Method \citep{nie2022diffusion} and Surrogate Process\citep{lee2023robust} on CIFAR-10. In particular, we implemented our standalone validation process in their original pipelines to validate their attacks and defenses. The results are summarized in \Cref{tab:table8}. In our validation pipeline, Diffpure Adjoint Attack has a high adversarial accuracy, showing its attack is actually not very strong, which may give a false sense of security for the associated defense method. Thus, Diffpure must be re-evaluated by a strong FULL gradient attack, which is what we have presented in \Cref{tab:table1} and stated the why it's important for the evaluation in \Cref{sec:attack_process}. On the other hand, Surrogate Process achieves much stronger (i.e., lower) adversarial accuracy than Diffpure Adjoint Method. However, it is still 8\% weaker (i.e., higher) than our MEFA framework on the same images. Note that the Surrogate Process starts with a lower natural accuracy due to the surrogate denoising steps. Overall, MEFA’s attack reduces much more accuracy during the attack process, thus a much stronger attack. As stated in the paper, we emphasize that MEFA has the state-of-the-art attack success rate on CIFAR-10, i.e., the largest accuracy reduction during the attack process. 

For runtime, Diffpure Adjoint Method appears to be even slower than our MEFA framework. It takes 18 minutes per image, whereas Diffpure takes 23 minutes per image on the same hardware. We understand that the runtime of the Diffpure Adjoint Method can be slightly improved with a larger batch size if memory allows; however, its runtime advantage will not be significant. Surrogate Attack takes about 4 minutes per image on the same hardware. We acknowledge Surrogate Attack’s speed advantage, yet we believe a stronger attack is important for defense evaluation, and the longer runtime of MEFA is worth the trade-off. Future accelerated full-gradient attacks may be inspired by MEFA and use it as a baseline.

\begin{table}[ht]
\centering
\caption{\small Score SDE Defense against PGD+EOT20 attacks from WRN-28-10 on CIFAR-10.}
\label{tab:table8}
\vspace{0.02in}
\resizebox{\textwidth}{!}{ % Resize table to fit page width 
\begin{small}
\begin{tabular}{p{5.5cm}ccccc}
\toprule
\textbf{Defense} & \textbf{Attack} & \multicolumn{2}{c}{\textbf{Nat. Accuracy (NA)}} & \multicolumn{2}{c}{\textbf{Adv. Accuracy (AA)}} \\
\cmidrule(lr){3-4} \cmidrule(lr){5-6}
\textbf{Defense Model} & \textbf{Type~~\,} & $H_{\mathrm{d}}=1$ & $H_{\mathrm{d}}=50$ & $H_{\mathrm{d}}=1$ & $H_{\mathrm{d}}=50$ \\
\midrule
{Diffpure Adjoint Method (n$^*$=100) } 
& $\ell_\infty$ & 93.33\% & 93.33\% &  88.33\% & {\bftab 91.67\%} \\
\midrule
{Surrogate Process (n$^*$=510)} 
& $\ell_\infty$ & 80.00\% & 87.06\% &  50.39\% & {\bftab 50.59\%} \\

\bottomrule
\footnotesize{$^*$n=number of random sampled images}
\end{tabular}
\end{small}
}
\end{table}

\section{Smooth and Non-Smooth EBM}\label[appendix]{sec:EBM_comparison}
The adaptive PGD attack is based on gradient update. \citep{athalye2018obfuscated} has pointed out the obfuscated gradients that cause false security. and proposed BPDA to circumvent the problem of exploding or vanishing gradients. However, the experiments show in \Cref{tab:table3} that BPDA does not solve this problem when the gradients become very small due to the many layers of the networks, even with LeakyRelu or SoftLeakyRelu shown in \Cref{tab:table4}. EBM sampling evolves gradient update, and thus during the backpropagation, the LeakyRelu second-order derivative is 0. Thus, we use SoftLeakyRelu with non-zero second-order derivative for PGD attack. We see that SoftLeakyRelu has a more stable gradient update and resulted in a stronger PGD attack. 

\subsection*{Non-Smooth EBM with LeakyRelu}
Recall the LeakyRelu and its gradient:
\begin{align*}
\text{LeakyRelu}(\mathbf{x}) &= 
\begin{cases} 
\mathbf{x}, & \text{if } \mathbf{x} > 0; \\
0.05\mathbf{x}, & \text{otherwise.}
\end{cases}\\
\frac{\partial}{\partial \mathbf{x}} \text{LeakyRelu}(\mathbf{x}) &= 
\begin{cases} 
1, & \text{if } \mathbf{x} > 0; \\
0.05, & \text{otherwise.}
\end{cases}
\end{align*}
For negative inputs (\(\mathbf{x} \leq 0\)), the gradient is scaled by \(\alpha\), a small value (e.g., \(0.05\)).
In \textbf{deep networks}, gradients for early layers are computed via the \textbf{chain rule}:
\begin{equation}    
    \frac{\partial L}{\partial w_{\text{layer}_1}} = \frac{\partial L}{\partial \alpha_{\text{last}}} \cdot \prod_{i=2}^{n} \alpha_{i}.
\end{equation}    
After \(n=5\) layers with LeakyRelu, the gradient decays exponentially approximately as $0.05^{5} = 3.125e^{-7}$. 

\subsection*{Smooth EBM with SoftLeakyRelu}
Recall SoftLeakyRelu:
\[
\text{SoftLeakyRelu}(\mathbf{x}) = 
\underbrace{(1 - a)\mathbf{x}}_{\text{linear term}} + 
\underbrace{a\sqrt{\mathbf{x}^2 + e^2}}_{\text{smooth term}} - 
\underbrace{ae}_{\text{constant offset}},
\]
where we take \(a = 0.49\) (controls the ``leakiness'') and \(e = 0.01\) (smoothing parameter). Thus, its gradient is:
\begin{equation}
\frac{\partial}{\partial \mathbf{x}} \text{SoftLeakyRelu}(\mathbf{x}) = 0.51 + \frac{0.49x}{\sqrt{\mathbf{x}^2 + 0.0001}}.
\end{equation}
For $\mathbf{x} \gg 0$, gradient approaches 1.0, for $\mathbf{x} \ll 0$, gradient approaches 0.02 and At $\mathbf{x}=0$, gradient is 0.51.

\begin{table}[ht]
\centering
\small{
\caption{EBM BPDA+EOT20 white-box attacks on CIFAR-10 with $\ell_\infty$ ($\varepsilon=8/255$).}
\label{tab:table4}
\resizebox{\textwidth}{!}{ % Resize table to fit page width
\begin{tabular}{p{5cm}lcccc}
\toprule
\textbf{Defense Model} & \textbf{Attack Method} & \multicolumn{2}{c}{\textbf{NA}} & \multicolumn{2}{c}{\textbf{AA}} \\
\cmidrule(lr){3-4} \cmidrule(lr){5-6}
 & & $H_{\mathrm{d}}=1$ & $H_{\mathrm{d}}=50$ & $H_{\mathrm{d}}=1$ & $H_{\mathrm{d}}=50$ \\
\midrule
\multirow{2}{*}{Smooth EBM} 
& adaptive step size & 75.88\% & 84.51\% & 56.67\% & 55.10\% \\
& fixed stepsize & 75.88\% & 84.51\% & 44.14\% & 49.80\% \\
\midrule
\multirow{2}{*}{Non-Smooth EBM} 
& adaptive step size & 70.00\% & 81.12\% & 55.29\% & 53.50\% \\
&  fixed step size* with $H_{\mathrm{d}}=150$ \citep{hill2021stochastic} & - & 84.12\% 
& - & 54.90\% \\
\bottomrule
\multicolumn{4}{l}{%
  \begin{minipage}{6.5cm}%
    \small *results from the paper.%
  \end{minipage}%
}\\
\end{tabular}
\vspace{0.2cm}

}
}
\end{table}

\end{document}